\pdfoutput=1
\documentclass[runningheads]{llncs}

\usepackage{eccv}

\usepackage{eccvabbrv}
\usepackage{graphicx}
\usepackage{booktabs}
\usepackage[accsupp]{axessibility}

\usepackage{hyperref}

\usepackage{orcidlink}

\usepackage{multirow}
\usepackage{amssymb}
\usepackage{xcolor}
\usepackage{float}
\usepackage{algorithm}
\usepackage{algorithmic}

\usepackage{listings}
\begin{document}

\title{FlashBEV: Fast and Memory-Efficient Exact BEV Transformation with IO-Awareness}
\titlerunning{FlashBEV: Fast and Memory-Efficient Exact BEV Transformation}
\author{%
Shunsuke Yokokawa\inst{1,2}\orcidlink{0009-0002-7990-6106} \and
Hironori Kasahara\inst{1}\orcidlink{0000-0001-7984-756X}}
\authorrunning{S. Yokokawa and H. Kasahara}
\institute{%
Waseda University, Tokyo, Japan\\
\email{yokosyun@fuji.waseda.jp, kasahara@waseda.jp} \and
T2, Inc., Tokyo, Japan}

\maketitle

\begin{abstract}
Bird's-eye-view (BEV) perception is a core component of camera-based 3D understanding in autonomous driving, where view transformation (VT) maps multi-camera image features into a unified BEV representation. Sampling-based view transformation (Sampling-VT) is attractive because it supports dense and continuous BEV aggregation for high-resolution and long-range perception. Its deployment bottleneck, however, is systems-level: standard tensorized implementations of Sampling-VT---which we refer to as Tensorized Sampling-VT---explicitly materialize large height-dependent intermediate tensors, causing memory and latency costs that scale poorly with vertical resolution and the number of cameras.

We revisit Tensorized Sampling-VT from an operator-execution perspective and show that it follows a gather--reduction pattern: each BEV query independently accumulates contributions across cameras and height bins. Unlike splatting-based VT, which requires index sorting and prevents fully thread-local reduction from voxel construction to BEV output, the gather--reduction structure of Sampling-VT enables thread-local accumulation with on-the-fly recomputation, eliminating the need to materialize height- and camera-dependent intermediates.

Based on this insight, we propose \textbf{FlashBEV}, a fully fused and IO-aware execution strategy that is mathematically equivalent to Tensorized Sampling-VT (same operator output) while substantially reducing global memory traffic and kernel-launch overhead. Experiments show that FlashBEV achieves more than an order of magnitude lower peak GPU memory and significant inference-latency speedups, with memory usage effectively independent of the number of height bins, reducing the operator's peak memory to $O(BCXY)$ (output only). This unlocks higher BEV range/resolution and vertical discretization within fixed deployment budgets on memory-constrained devices, where tensorized execution would otherwise be infeasible. Our contribution is therefore an execution redesign---same math, different execution---that removes a key scalability barrier for deployment-ready Sampling-VT. Code is available at \url{https://github.com/yokosyun/FlashBEV}.

\end{abstract}

\section{Introduction}
\label{sec:intro}
Bird's-eye-view (BEV) perception is a fundamental representation for camera-based 3D scene understanding in autonomous driving. A central operator in BEV pipelines is \emph{view transformation} (VT), which maps image-space features into a shared BEV coordinate frame. Because VT is executed on every frame, its computational efficiency directly affects end-to-end deployability.

\begin{figure}[t]
    \centering
    \includegraphics[width=\linewidth]{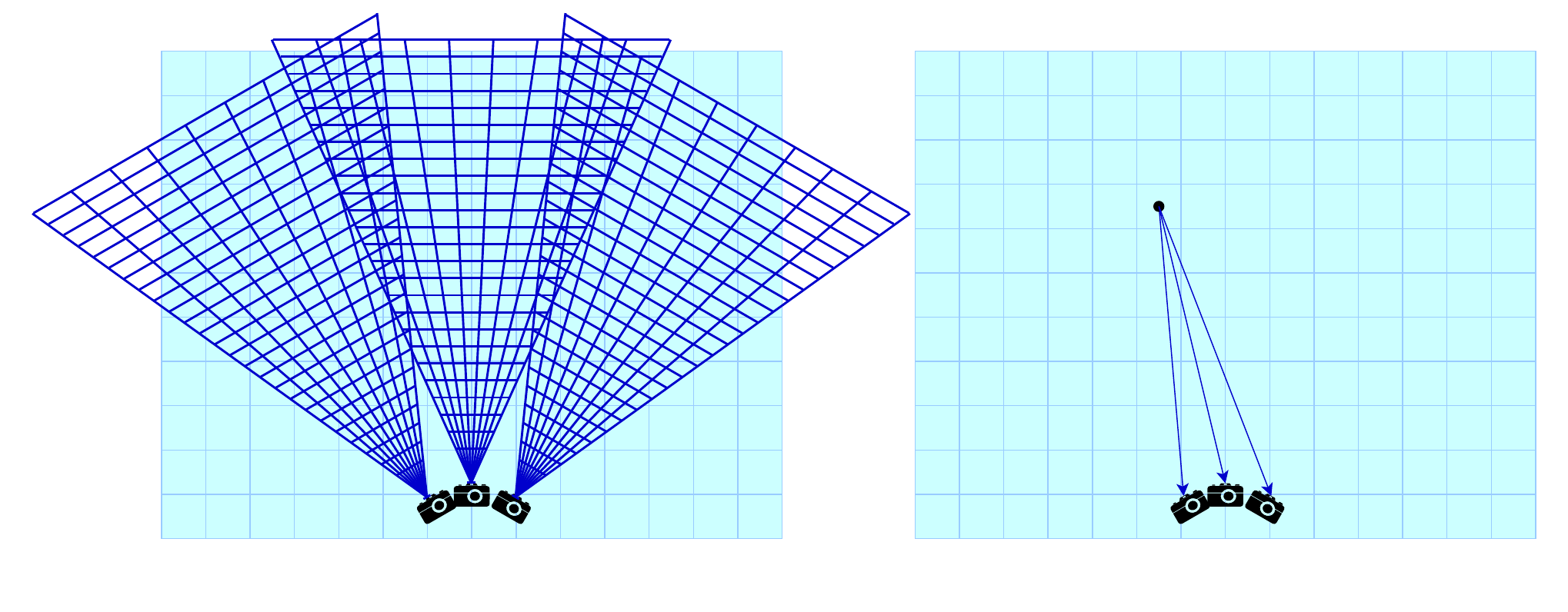}
    \caption{Conceptual comparison of Splatting-VT (left), which forward-projects (scatters) image features into BEV, and Sampling-VT (right), which backward-queries (gathers) features at each BEV location.}
    \label{fig:splatting-vt_vs_sampling-vt}
\end{figure}

Existing VT designs are broadly divided into \emph{splatting-based VT (Splatting-VT)} and \emph{sampling-based VT (Sampling-VT)}. As shown in Fig.~\ref{fig:splatting-vt_vs_sampling-vt}, Splatting-VT forward-projects image features into BEV and is often memory-friendly with index-based pooling~\cite{bevpoolv2}, but it tends to produce geometry-dependent, uneven coverage. Sampling-VT instead performs backward querying from each BEV location into image space, enabling dense and continuous aggregation over the BEV grid. This density makes Sampling-VT attractive for high-resolution and long-range perception.

The challenge is that Sampling-VT is expensive under standard tensorized execution. In this commonly used baseline (\emph{Tensorized Sampling-VT}, i.e., tensorized gather with height-sum reduction), large intermediate tensors are explicitly materialized over BEV cells, cameras, and height bins. Under representative settings (e.g., 50\,m BEV range, grid $200\times200$, 8 height bins, batch size 1), the view-transformation module alone can reach nearly 2\,GB peak memory for the forward pass; longer range or higher resolution further increases memory and latency. In practice, this memory-traffic bottleneck, rather than arithmetic throughput, limits deployment.
Moreover, while view transformation may not dominate end-to-end latency at standard resolutions, it becomes a primary bottleneck---or triggers out-of-memory failures---as BEV range, BEV resolution, or vertical discretization are scaled.

In this paper, we argue that the bottleneck is not the math of Sampling-VT, but its \emph{execution}. We analyze Tensorized Sampling-VT at operator level and observe a gather--reduction structure: each BEV query can independently accumulate contributions across cameras and height bins, so the height-expanded tensor $\mathbb{R}^{B\times N\times C\times X\times Y\times Z}$ need not be materialized---sampling and reduction fuse into a thread-local accumulation that directly produces the BEV feature map.

\paragraph{Contributions.}
Our contribution is an \emph{execution design}, not a new learning algorithm: we preserve the exact Tensorized Sampling-VT operator while changing how it runs on GPU. Specifically, we make the following contributions:

\begin{itemize}
    \item We identify Tensorized Sampling-VT as a \emph{gather--reduction} computation and expose the source of its memory-traffic inefficiency in tensorized execution.

    \item We derive a \emph{single fused execution} that performs sampling and accumulation with thread-local reduction and on-the-fly recomputation, eliminating any intermediate tensors.

    \item We propose \textbf{FlashBEV}, an IO-aware implementation of this execution that remains \emph{exactly equivalent} to Tensorized Sampling-VT while reducing memory traffic and kernel overhead.

    \item We show through experiments that FlashBEV delivers large gains in efficiency---more than an order of magnitude lower peak memory and substantial latency speedups---with memory usage effectively independent of the number of height bins.
\end{itemize}

To the best of our knowledge, FlashBEV is the first method that executes dense Sampling-VT without materializing camera- and height-dependent tensors and without relying on lookup tables; by removing this fundamental memory bottleneck, FlashBEV opens a new design space for scalable BEV perception and future algorithmic exploration. FlashBEV is orthogonal to model design and can replace dense Sampling-VT in existing pipelines (e.g., SimpleBEV); extension to multi-resolution gathering is future work.

\section{Related Work}
\label{sec:related_work}
\subsection{Splatting vs.\ Sampling}
We distinguish \emph{Splatting-VT} (scatter) \cite{lss, bevsan, BEVFusionMIT, bevpoolv2, CaDDN, GaussianLSS, bevdet, bevdet4d, bevdepth, bevheight, ealss, lssinst, rayformer} from \emph{Sampling-VT} (gather) \cite{simplebev, bevformer, bevformerv2, polarformer, fastbev, fastbevpp, pointbev, heightformer, parametricbev, hvbev}. Tensorized implementations of Sampling-VT materialize height-dependent intermediate tensors; to our knowledge, no prior work removes this tensor while preserving exact Tensorized Sampling-VT (gather) semantics.

\subsection{Sparse Sampling}
\label{sec:reduce_sampling_counts}

\noindent Several works reduce the computational cost of Sampling-VT by evaluating only a subset of spatial or vertical samples instead of performing dense aggregation. These strategies include restricting computation to selected BEV regions, coarse-to-fine or adaptive sampling schemes, and learned sparse sampling mechanisms such as deformable attention that predict sampling offsets~\cite{deformable_detr}.

\textbf{Sparse BEV Sampling.}
A number of works reduce the cost of dense BEV querying by activating only a subset of BEV cells or refining selected regions. PointBeV~\cite{pointbev} adopts a coarse-to-fine strategy and activates only a subset of BEV cells, reducing computation and memory compared to dense grids. FB-BEV~\cite{fbbev} refines only foreground BEV regions by combining forward and backward view transformations, reducing redundant computation over background areas. SparseBEV~\cite{sparsebev} avoids dense BEV construction entirely via a fully sparse, query-based detector with adaptive spatio-temporal sampling from multi-camera features.

\textbf{Sparse Height Sampling.}
Another line of work reduces the cost of Sampling-VT by decreasing the number of height samples (or 3D points) evaluated per BEV location. Instead of uniformly discretizing the vertical dimension, these approaches adaptively allocate height samples according to geometric priors or object distributions. HV-BEV~\cite{hvbev} further introduces height-aware reference point sampling, decoupling horizontal feature aggregation and vertical sampling so that the number of sampled heights better matches object height distributions in the scene. These approaches reduce computation by sparsifying vertical sampling, but they still aggregate over multiple height samples or learned offsets per BEV location (e.g., deformable attention over offsets and multi-scale features), and typically maintain intermediate tensors that incur additional memory cost under tensorized execution when full-resolution aggregation is required.

Both lines of work trade full dense BEV coverage (or dense vertical sampling) for sparsification and do not address the \emph{implementation-level} inefficiency of Tensorized Sampling-VT when full-resolution aggregation is required.

\subsection{Lookup-Table Based Sampling-VT}
\label{sec:lut_based_sampling_vt}
An alternative to dense tensorized execution is to precompute geometric correspondences in a lookup table (LUT) and replace runtime projection with indexed gathering; this reduces compute and memory but typically relies on discretized image coordinates and thus does not preserve exact continuous Sampling-VT semantics. Several works adopt LUT-based projection to this end and significantly accelerate BEV feature construction. LUT-based methods differ along two dimensions: when the LUT is built (online vs.\ offline) and how geometry is used for aggregation (voxel-based vs.\ interval-based).

\textbf{Online vs.\ offline LUTs.}
Online LUTs are built at runtime and adapt to camera changes but incur per-frame construction and validity checks. Offline LUTs assume static cameras and avoid per-frame cost but require rebuilding when parameters change.

\textbf{Voxel-based LUT projection.}
FastBEV~\cite{fastbev} (typically with offline precomputation) precomputes a mapping from each BEV voxel to a corresponding image pixel (single-hit per voxel) and performs feature aggregation via indexed gathering. To simplify the projection, each voxel is associated with a single camera source, so it can only sample from a single camera's features even when a BEV location is viewed by multiple cameras. While this design reduces computation and memory access, it limits multi-view fusion in overlapping or partially occluded regions. In addition, FastBEV constructs intermediate 3D voxel features before collapsing them to BEV, leading to memory usage that scales with the number of height bins.

\textbf{Interval-based LUT projection.}
Other approaches adopt interval-based indexing inspired by BEV pooling operators. For example, DualBEV~\cite{dualbev} (often with offline precomputation) aggregates image features into BEV grids using precomputed intervals that specify which pixels contribute to each BEV location. This strategy avoids materializing intermediate 3D voxel volumes and improves memory efficiency compared to the voxel-based LUT approach above. However, it still requires additional indexing structures to store image indices and interval ranges, and uses discretized pixel indices.

Despite their efficiency benefits, LUT-based approaches rely on discretized image coordinates obtained through rounding or quantization of projected pixel locations. This discretization can introduce approximation errors in feature sampling and requires constructing LUT structures that depend on camera parameters. FlashBEV, in contrast, preserves exact continuous Sampling-VT semantics (bilinear interpolation and camera aggregation) with no precomputed LUT and no height-dependent intermediates, using only the same inputs as the tensorized baseline: camera projection matrices $(B,N,3,4)$ and image feature maps. It requires no indexing structures that scale with the BEV grid or voxel space.

\subsection{Kernel Fusion}
Basic kernel fusion is typically limited to element-wise operators (e.g., ReLU, Mul, Add) or short predefined patterns in deployment frameworks such as TensorRT~\cite{tensorrt} or TVM~\cite{tvm}. FlashAttention~\cite{flashattention} shows that reordering computation and selectively recomputing intermediates can \emph{enable} much more aggressive kernel fusion, substantially reducing HBM reads/writes while preserving exact semantics. FlashBEV applies the same idea to Sampling-VT: by reordering voxel construction, voxel projection, and validity handling, it achieves a single fused kernel with thread-local accumulation and no height-dependent intermediate tensors, while remaining exactly equivalent to Tensorized Sampling-VT and using minimal HBM (inputs and output only). FlashBEV thus addresses the gap left by sparsification and LUT-based methods by enabling fully fused execution (Sec.~\ref{sec:method}) without sparsification or discretization.

\section{Method}
\label{sec:method}
\begin{figure}[t]
  \centering
  \includegraphics[width=\linewidth]{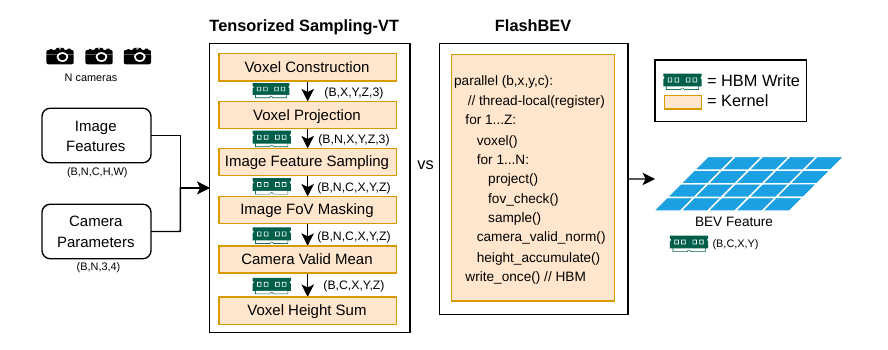}
  \caption{Left: Tensorized Sampling-VT materializes multiple intermediate tensors in HBM; computation and memory scale as $O(BNCXYZ)$. Right: FlashBEV assigns one thread per $(b,x,y,c)$, performs voxel construction, projection, masking, feature sampling, and reduction locally, and writes only the final BEV feature to global memory.}
  \label{fig:flashbev_overview}
\end{figure}

We first formalize Tensorized Sampling-VT and its design choices, then analyze the
intrinsic memory bottleneck under fully tensorized execution. Based on these
observations, we propose \textbf{FlashBEV}, a reordered execution strategy that
eliminates intermediate tensor materialization while preserving exact
equivalence to the Tensorized Sampling-VT formulation. Fig.~\ref{fig:flashbev_overview} gives an overview of the contrast between the baseline and our approach.

\subsection{Preliminary}
\label{subsec:preliminary}

We consider a batch size $B$, $N$ cameras, a BEV grid of size $(X,Y)$, $Z$ height
bins, and $C$ feature channels. Sampling-VT aggregates camera features into a
BEV feature map $B(x,y,c)$ by computing the \emph{camera-wise valid mean} at each
height bin $z$, then summing over $z$:
\begin{equation}
B(x,y,c)=
\sum_{z=1}^{Z}
\frac{
\sum_{n=1}^{N} M_n(x,y,z)\, f_n(x,y,z,c)
}{
\max\!\left(1,\sum_{n=1}^{N} M_n(x,y,z)\right)
}.
\label{eq:sampling_vt_agg}
\end{equation}
Here, $f_n(x,y,z,c)$ is the sampled image feature at the projected location,
and $M_n(x,y,z)\in\{0,1\}$ indicates whether voxel $(x,y,z)$ projects to a valid
image location for camera $n$.
The mask $M_n$ is 1 when the projected voxel center lies within the camera image with positive depth, and 0 otherwise; $f_n$ is bilinear interpolation from the camera feature map at the projected pixel.

\subsubsection{Tensorized Sampling-VT: Baseline}
\label{subsec:tensorized_execution}

We adopt the canonical Sampling-VT formulation (Eq.~\eqref{eq:sampling_vt_agg}) and implement it using a fully tensorized execution that explicitly materializes height-dependent intermediate tensors. We refer to this baseline implementation as \emph{Tensorized Sampling-VT}. Prior pipelines (e.g., SimpleBEV) may differ in reduction or implementation details; we use this generic baseline so that efficiency gains are measured consistently. Sec.~\ref{subsec:baseline_validation} shows that replacing height conv with height sum incurs only a small accuracy drop. Both the baseline and FlashBEV realize Eq.~\eqref{eq:sampling_vt_agg} by computing the valid mean over cameras at each height bin $z$, then summing over $z$. Algorithm~\ref{alg:tensorized_vt} summarizes the baseline execution and its HBM traffic.

\subsection{Observations}
\label{subsec:observations}

\subsubsection{Intrinsic Memory Bottleneck under Tensorized Execution}
\label{subsec:memory_bottleneck}

Under tensorized execution, Sampling-VT materializes sampled 3D tensors whose shapes
scale with the height discretization $Z$, e.g.,
\begin{equation}
\texttt{Features3D}\in\mathbb{R}^{B\times N\times C\times X\times Y\times Z}.
\end{equation}
These intermediates dominate peak memory and HBM traffic; memory grows linearly with $Z$, a primary scalability bottleneck. Computation and memory scale as
$O(BNCXYZ)$.

\begin{algorithm}[t]
\caption{Tensorized Sampling-VT (Baseline)}
\label{alg:tensorized_vt}
\begin{algorithmic}
\REQUIRE Camera features $\mathbf{F}_n$ ($N$ cameras), projection matrices $\mathbf{P} \in \mathbb{R}^{B\times N\times 3\times 4}$; all in HBM. BEV grid $(X,Y)$, height bins $Z$, channels $C$.
\STATE \textbf{Voxel construction:} build voxel grid; shape $(B,X,Y,Z,3)$. \textbf{Write} voxel grid to HBM.
\STATE \textbf{Voxel projection:} project each voxel to each camera; shape $(B,N,X,Y,Z,3)$. \textbf{Write} projected coords to HBM.
\STATE \textbf{Image feature sampling:} bilinear sample $\mathbf{F}_n$ at projected locations; Features3D $\in \mathbb{R}^{B\times N\times C\times X\times Y\times Z}$. \textbf{Write Features3D to HBM.}
\STATE \textbf{FoV masking:} compute validity $M_n(x,y,z) \in \{0,1\}$ from projected coords; apply mask to Features3D (element-wise). \textbf{Write} $M_n$ and masked Features3D to HBM.
\STATE \textbf{Valid camera mean:} for each $(x,y,z)$ compute $G(x,y,z,c) = \mathrm{num}_z/\max(1,\mathrm{den}_z)$ with $\mathrm{num}_z = \sum_n M_n f_n$, $\mathrm{den}_z = \sum_n M_n$. \textbf{Write $G$ to HBM.}
\STATE \textbf{Voxel height sum:} $B(x,y,c) \leftarrow \sum_z G(x,y,z,c)$. \textbf{Write $B$ to HBM.}
\STATE \textbf{return} $B$
\end{algorithmic}
\end{algorithm}

\subsubsection{Sampling-VT as a Gather--Reduction Operator}
\label{subsec:gather_observation}

Despite the memory cost of tensorized execution, Eq.~\eqref{eq:sampling_vt_agg} reveals
a simple structure: for each BEV location $(x,y)$ and channel $c$, the output is
a sum over height bins $z$, where at each $z$ a gather--reduction over cameras $n$
yields the valid mean (numerator $\mathrm{num}_z=\sum_n M_n f_n$, denominator
$\mathrm{den}_z=\sum_n M_n$, then $\mathrm{num}_z/\max(1,\mathrm{den}_z)$), and
$B=\sum_z$ of that. Thus Sampling-VT is a \emph{gather--reduction} operator with
per-height-bin normalization. This computation has no cross-$(x,y)$ dependencies,
suggesting that each output BEV element can be computed independently using
thread-local accumulation. This gather--reduction structure implies that
Sampling-VT can be executed as a single fused operator, computing each BEV element
independently without staging intermediate tensors.

Splatting-VT requires index sorting and blocks fully thread-local reduction; Sampling-VT's gather pattern admits reordering so each BEV element can be computed independently. We therefore design FlashBEV to exploit this structure; it uses minimal HBM (inputs and output only).

\subsection{FlashBEV}
\label{subsec:flashbev}

\subsubsection{Thread-Local Execution and Reduction}
\label{subsec:thread_local}

Fig.~\ref{fig:flashbev_overview} contrasts FlashBEV with Tensorized
Sampling-VT. Tensorized execution materializes multiple intermediate tensors in HBM and performs global reductions over $(n,z)$ via global memory. 
In contrast, FlashBEV launches a fused CUDA kernel over output indices $(b,x,y,c)$, where each thread computes a single BEV element by performing a local reduction over cameras $n$ and height
bins $z$, and writes the result once to global memory.

Concretely, each GPU thread is assigned a BEV location $(x,y)$ and channel $c$.
The thread iterates over height bins $z$; for each $z$, it iterates over cameras $n$,
performs projection, validity checking, and bilinear sampling on-the-fly, and accumulates the per-$z$ numerator and denominator
($\mathrm{num}_z\leftarrow \mathrm{num}_z + M_n\, f_n$,
$\mathrm{den}_z\leftarrow \mathrm{den}_z + M_n$).
After completing the loop over $n$, the thread sets
$G(x,y,z,c)=\mathrm{num}_z/\max(1,\mathrm{den}_z)$. After looping over all $z$, it writes
$B(x,y,c)=\sum_z G(x,y,z,c)$ to HBM. All intermediate quantities exist only
transiently in registers and are discarded immediately
after use, eliminating intermediate tensor materialization. Algorithm~\ref{alg:flashbev} gives the corresponding pseudocode.

\begin{algorithm}[t]
\caption{FlashBEV}
\label{alg:flashbev}
\begin{algorithmic}
\REQUIRE Camera features $\mathbf{F}_n$, projection matrices $\mathbf{P}$; in HBM. Output $B \in \mathbb{R}^{B\times X\times Y\times C}$ in HBM (initialized to zero).
\STATE One thread per output $(b,x,y,c)$. All per-thread data kept in registers.
\STATE Initialize $\mathrm{acc} \leftarrow 0$ (register).
\FOR{$z = 1, \ldots, Z$}
\STATE $\mathrm{num}_z \leftarrow 0$, $\mathrm{den}_z \leftarrow 0$
\FOR{$n = 1, \ldots, N$}
\STATE Compute voxel center; project to camera $n$
\STATE If not valid: \textbf{continue}
\STATE $f_n \leftarrow \mathrm{bilinear}(\mathbf{F}_n)$
\STATE $\mathrm{num}_z \leftarrow \mathrm{num}_z + f_n$
\STATE $\mathrm{den}_z \leftarrow \mathrm{den}_z + 1$
\ENDFOR
\STATE $g_z \leftarrow \mathrm{num}_z / \max(1, \mathrm{den}_z)$
\STATE $\mathrm{acc} \leftarrow \mathrm{acc} + g_z$
\ENDFOR
\STATE \textbf{Write} $B(b,x,y,c) \leftarrow \mathrm{acc}$ to HBM once
\STATE \textbf{return} $B$
\end{algorithmic}
\end{algorithm}

\noindent

Peak memory is $O(BCXY)$, independent of $Z$ and $N$ (Fig.~\ref{fig:flashbev_overview}).

\subsubsection{Operator-Level Equivalence}
\label{subsec:operator_equivalence}

FlashBEV is mathematically equivalent to Tensorized Sampling-VT: the same interpolation and reduction (Eq.~\eqref{eq:sampling_vt_agg}) are applied; only the execution order and storage of intermediates change. Inference matches up to floating-point rounding (max error on the order of $10^{-4}$, Sec.~\ref{subsec:baseline_validation}); training with FlashBEV yields the same or comparable convergence (Sec.~\ref{subsec:baseline_validation}).

\subsubsection{Memory Computation Trade-Off}
\label{subsec:tradeoff}
FlashBEV trades increased arithmetic for reduced memory traffic: channel-independent quantities (voxels, projection, validity) are recomputed per channel instead of stored, avoiding $O(BNCXYZ)$ HBM traffic and shifting execution toward compute-bound.

\section{Experiments}
\label{sec:experiments}

We evaluate numerical correctness, efficiency, and scalability at the view-transform level under identical network architectures and configurations.

\subsection{Experimental Setup}
\label{subsec:exp_setup}

\paragraph{Reference models.}
We use the SimpleBEV pipeline as the reference BEV architecture: same spatial range and voxelization ($x,y \in [-50,50]$\,m at 0.5\,m resolution; $z \in [-5,5]$\,m at 1.25\,m resolution), input image size $448\times800$, backbone downsampling factor 8 (feature map $56\times100$ per camera), and $C=128$ channels. Unless otherwise stated, batch size is 1 and the default BEV grid is $X=Y=200$ with $Z=8$ height bins; we vary $Z$, grid size, or $C$ in scaling experiments. Only the view transformation module is replaced by FlashBEV; all other components are unchanged. For detection (baseline validation and E2E) we follow the SimpleBEV training and evaluation protocol; for efficiency we benchmark the view transformation module independently. Tensorized Sampling-VT (our baseline) is implemented as a fully vectorized PyTorch module that materializes the intermediate voxel feature tensor of shape $B\times N\times C\times X\times Y\times Z$; FlashBEV uses fused CUDA kernels and accumulates directly into the BEV grid without intermediate 3D tensors. Under identical configurations both produce numerically equivalent BEV feature maps.

\paragraph{Metrics and protocol.}
We report inference latency, peak GPU memory, and numerical differences at the BEV feature level. Detection accuracy (IoU) is reported as mean $\pm$ standard deviation over 5 random seeds. All experiments use FP32; we time 100 runs after warm-up with \texttt{torch.cuda.synchronize()} and report mean latency; peak memory is \texttt{torch.cuda.max\_memory\_allocated} during the VT forward pass. Profiling on NVIDIA RTX A6000 (multi-GPU results in Fig.~\ref{fig:latency_speedup_gpus}); training on 8$\times$ NVIDIA H200.

\subsection{Accuracy Analysis}
\label{subsec:baseline_validation}

We validate correctness at both task level (detection IoU) and operator level (BEV feature and gradient error). Table~\ref{tab:conv_vs_sum} reports detection IoU on nuScenes~\cite{Caesar2019nuScenesAM}. We first reproduce the SimpleBEV baseline (M1, materialized) and implement Tensorized Sampling-VT (M2, materialized) and FlashBEV (M3, fused) within the same pipeline. M2 matches M1 in IoU (46.9 vs.\ 46.9); M3 (FlashBEV) matches M2 (46.8 vs.\ 46.9 within variance), preserving detection accuracy with only the execution strategy changed.

\begin{table}[H]
\centering
\caption{Accuracy analysis: height reduction and reimplementation (M1, M2), and FlashBEV (M3). Mean $\pm$ std over 5 seeds.}
\label{tab:conv_vs_sum}
\begin{tabular}{clllc}
\toprule
ID & Variant & Implementation & Height reduction & IoU \\
\midrule
M1 & SimpleBEV & Materialized & Conv2d($H{\cdot}C {\to} C$) & \textbf{46.9 $\pm$ 0.1} \\
M2 & Tensorized Sampling-VT & Materialized & Sum + Conv2d($C {\to} C$) & \textbf{46.9 $\pm$ 0.3} \\
M3 & FlashBEV (Ours) & Fused & Sum + Conv2d($C {\to} C$) & \textbf{46.8 $\pm$ 0.1} \\
\bottomrule
\end{tabular}
\end{table}

\paragraph{Operator-level equivalence.}
Table~\ref{tab:numerical_diff} reports negligible BEV and gradient errors (mean $\sim 10^{-6}$); Table~\ref{tab:conv_vs_sum} (M3) confirms equivalent detection accuracy.

\begin{table}[H]
\centering
\caption{View transformation (VT) BEV feature and gradient differences; errors negligible (floating-point rounding).}
\label{tab:numerical_diff}
\begin{tabular}{lcccc}
\toprule
Comparison & Max Error & Mean Error & Mean Rel.\ Error & Cosine Sim. \\
\midrule
Forward & $2.93\times10^{-4}$ & $8.79\times10^{-6}$ & $7.18\times10^{-5}$ & 1.00 \\
Backward & $1.83\times10^{-4}$ & $7.98\times10^{-6}$ & $9.22\times10^{-6}$ & 1.00 \\
\bottomrule
\end{tabular}
\end{table}

\subsection{Efficiency on Reference Config}
\label{subsec:efficiency_ref_config}

We use the reference config from Sec.~\ref{subsec:exp_setup} ($X{=}Y{=}200$, $Z{=}8$, $C{=}128$, batch size 1).

\paragraph{E2E performance.}
We report end-to-end peak memory and latency (Table~\ref{tab:e2e_performance}); peak memory decreases from $2211.3$\,MB to $876.8$\,MB ($\approx 2.5\times$ lower) and latency from $67.2$\,ms to $59.9$\,ms ($-7.3$\,ms, $\approx 11\%$ faster), while detection accuracy remains equivalent. In this reference configuration VT is not the dominant E2E latency component, and we do not decompose per-module latency bottlenecks; the main practical benefit of FlashBEV is \emph{scalability}: larger BEV grids and higher vertical discretization within fixed deployment budgets (Sec.~\ref{subsec:vt_scaling}).

\begin{table}[H]
\centering
\caption{End-to-end peak memory and latency for different view transformation (VT) modules. Same architecture and training; only the VT module differs.}
\label{tab:e2e_performance}
\begin{tabular}{lcc}
\toprule
Method & E2E Peak Mem (MB) & E2E Latency (ms) \\
\midrule
Tensorized Sampling-VT & 2211.29 & 67.19 \\
FlashBEV (Ours) & 876.76 & 59.91 \\
\bottomrule
\end{tabular}
\end{table}

\paragraph{VT forward and backward performance.}
Table~\ref{tab:vt_forward_backward} reports VT forward-only and backward-only latency and peak memory: FlashBEV reduces VT peak memory from $1971.9$\,MB to $52.8$\,MB ($\approx 37\times$ lower) and achieves about $5.2\times$ forward speedup over Tensorized Sampling-VT.

\begin{table}[H]
\centering
\caption{VT forward and backward latency and peak memory (reference config: $X{=}Y{=}200$, $Z{=}8$, $C{=}128$).}
\label{tab:vt_forward_backward}
\begin{tabular}{lcc}
\toprule
Method & Forward (ms / MB) & Backward (ms / MB) \\
\midrule
Tensorized Sampling-VT & $9.18 \pm 0.23$ / 1971.94 & $17.90 \pm 0.20$ / 1174.86 \\
FlashBEV (Ours) & $1.77 \pm 0.03$ / 52.82 & $3.27 \pm 0.07$ / 105.63 \\
\bottomrule
\end{tabular}
\end{table}

\begin{figure}[t]
  \centering
  \includegraphics[width=\linewidth]{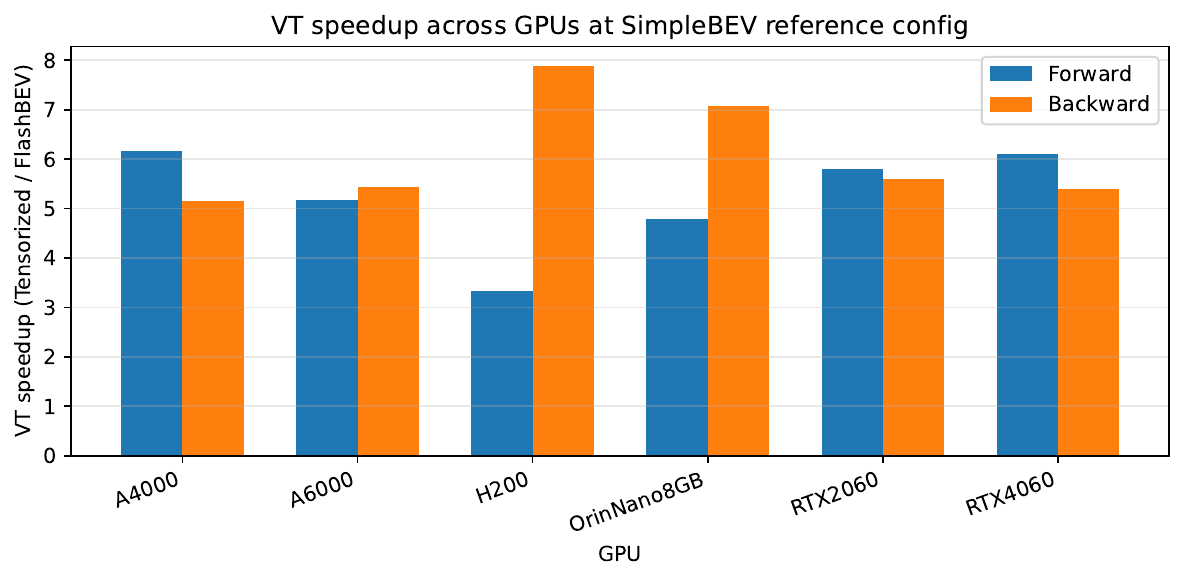}
  \caption{VT speedup (Tensorized Sampling-VT / FlashBEV) at SimpleBEV reference config ($X{=}Y{=}200$, $Z{=}8$, $C{=}128$) across six GPUs. FlashBEV achieves forward speedup 3.3--6.2$\times$ and backward 5.2--7.9$\times$ on all devices.}
  \label{fig:latency_speedup_gpus}
\end{figure}

\paragraph{VT speedup across GPUs.}
To show that gains are not specific to a single GPU, we evaluate VT speedup over Tensorized Sampling-VT on six devices: NVIDIA A4000, A6000, H200, Jetson Orin Nano 8GB, RTX 2060, and RTX 4060, including memory-constrained edge devices (e.g., Jetson Orin Nano 8GB). Fig.~\ref{fig:latency_speedup_gpus} reports forward and backward speedup. Across the six GPUs, forward speedup ranges from 3.3$\times$ to 6.2$\times$ and backward from 5.2$\times$ to 7.9$\times$.

\subsection{VT Scaling}
\label{subsec:vt_scaling}
We analyze how VT memory and latency scale when moving away from the reference config along key design axes (height resolution, BEV grid size, and channel count).

\paragraph{VT deployment capacity under fixed budgets.}
We set each budget to Tensorized Sampling-VT's cost at the reference config. Table~\ref{tab:deployment_capacity} gives the maximum $Z$ and maximum BEV grid $X{=}Y$ that FlashBEV achieves within that budget; under fixed memory, max $Z$ is N/A because FlashBEV's memory does not scale with $Z$.

\begin{table}[H]
\centering
\caption{Deployment capacity under fixed budgets (latency 9.18\,ms or memory 1972\,MB at reference config). Max $Z$ is N/A under fixed memory because FlashBEV's memory is independent of $Z$.}
\label{tab:deployment_capacity}
\begin{tabular}{lcc}
\toprule
Budget & Max $Z$ & Max BEV grid $X{=}Y$ \\
\midrule
Fixed latency (9.18\,ms) & $58$ (7.25$\times$) & $529$ (2.65$\times$) \\
Fixed memory (1972\,MB) & N/A & $1992$ (10$\times$) \\
\bottomrule
\end{tabular}
\end{table}

\begin{figure}[t]
  \centering
  \includegraphics[width=\linewidth]{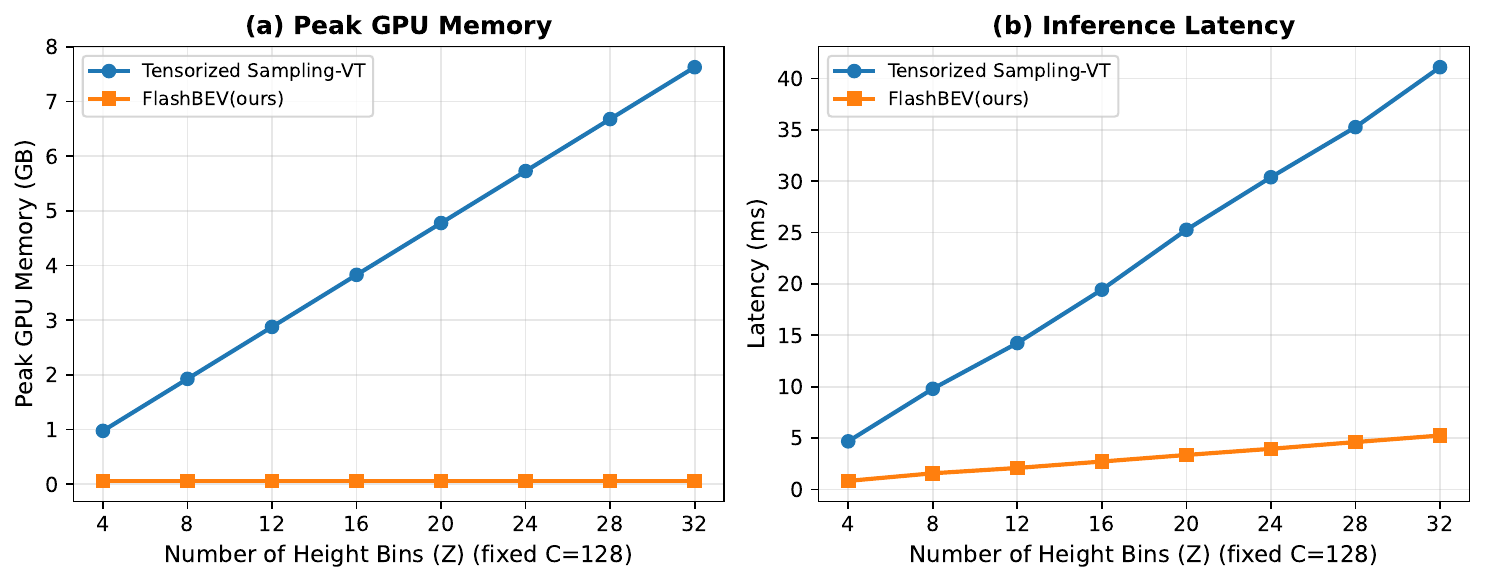}
  \caption{(a)~Peak GPU memory and (b)~inference latency vs.\ number of height bins (Z) with fixed BEV resolution and channels (reference config). Tensorized Sampling-VT exhibits peak memory and latency that grow roughly linearly with Z; FlashBEV achieves substantially lower memory and latency with weak dependence on Z by fusing sampling and reduction into a single kernel.}
  \label{fig:mem_lat_heightbins}
\end{figure}

\paragraph{VT memory and latency vs.\ height bins.}
As shown in Fig.~\ref{fig:mem_lat_heightbins}, Tensorized Sampling-VT exhibits near-linear growth in peak memory usage and latency as $Z$ increases, due to explicit materialization of intermediate 3D tensors whose size scales with $O(XYZ)$. FlashBEV maintains a substantially lower memory footprint that is independent of $Z$, and a much lower latency that grows far more slowly with $Z$ than the tensorized baseline, since it avoids height-dependent intermediates and accumulates within a fused kernel.

\begin{figure}[t]
  \centering
  \includegraphics[width=\linewidth]{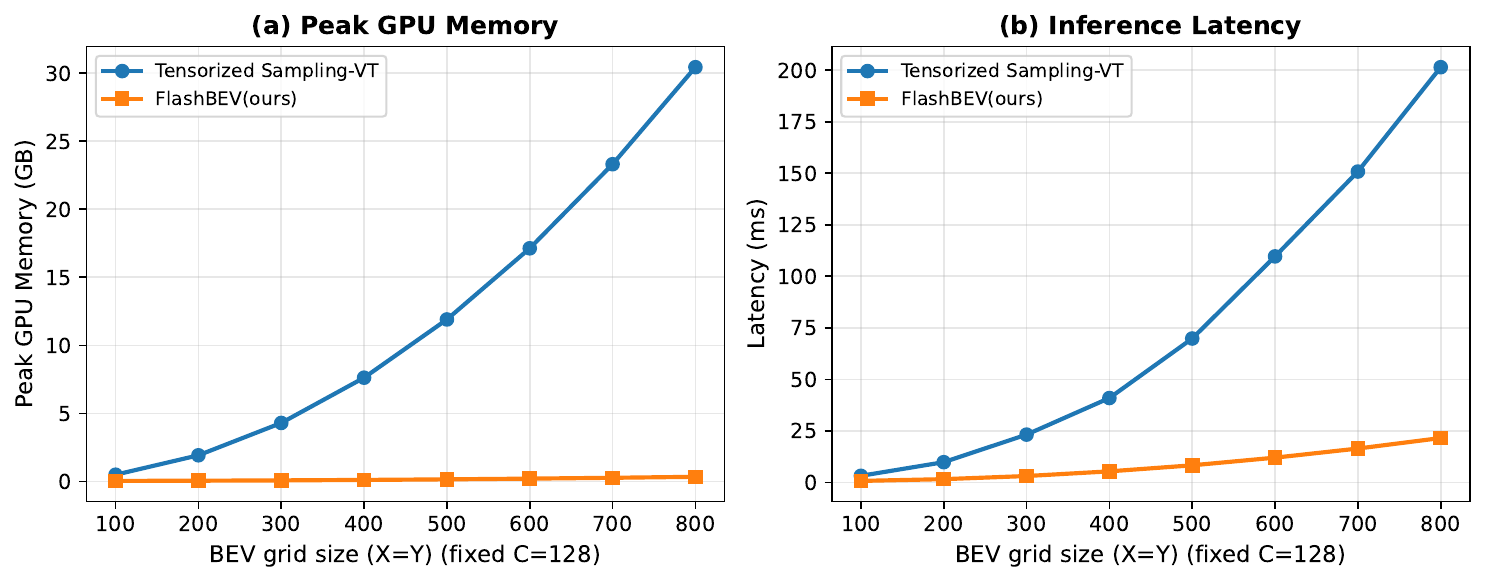}
  \caption{(a)~Peak GPU memory and (b)~inference latency vs.\ BEV grid size ($X{=}Y$) with fixed height bins and channels (reference config). Tensorized Sampling-VT exhibits strong growth in memory and latency with grid size; FlashBEV maintains much lower memory and latency, enabling larger grids under the same budget.}
  \label{fig:mem_lat_bev_grid}
\end{figure}

\paragraph{BEV grid scaling.}
We vary the BEV grid size $(X,Y)$ (longer range or finer resolution) and report the maximum grid achievable under fixed deployment constraints. Fig.~\ref{fig:mem_lat_bev_grid} shows how memory and latency scale with BEV grid size; Table~\ref{tab:deployment_capacity} gives deployment capacity under both budget types.

\begin{figure}[t]
  \centering
  \includegraphics[width=\linewidth]{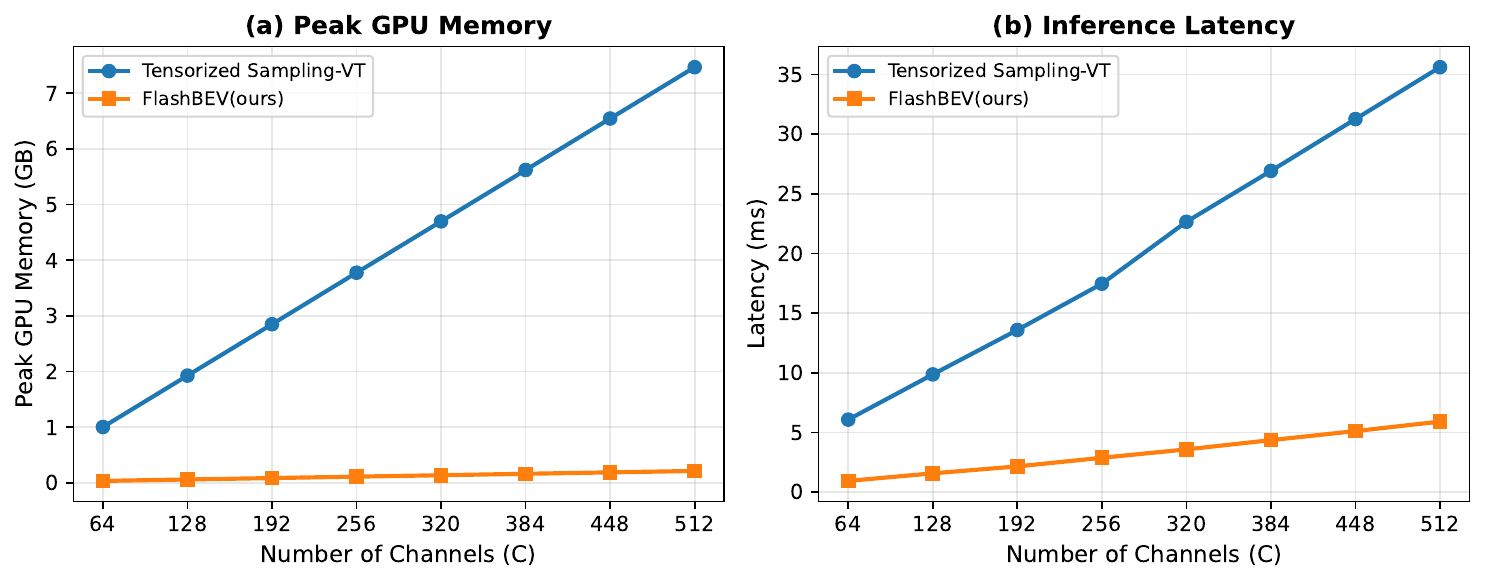}
  \caption{(a)~Peak GPU memory and (b)~inference latency vs.\ number of channels (C); reference config. FlashBEV achieves lower memory and high speedup for typical C; for very large C speedup can narrow but FlashBEV remains faster with much lower memory.}
  \label{fig:latency_channels}
\end{figure}

\paragraph{Channel dimension scaling.}
FlashBEV trades increased arithmetic for reduced memory traffic: channel-independent quantities (voxels, projection, validity) are recomputed per channel rather than stored as large tensors. Fixing the BEV grid ($X{=}Y{=}200$) and height bins ($Z{=}8$), we sweep $C$ (Fig.~\ref{fig:latency_channels}). For typical $C$ (128--256) FlashBEV keeps a substantial speedup; for larger $C$ recomputation cost grows and the relative speedup can narrow, but FlashBEV still avoids the OOM failures of the baseline.

\subsection{Deployment in TensorRT and Cross-Framework Applicability}
\label{subsec:tensorrt_crossframework}
The preceding results use SimpleBEV in PyTorch; we now show the same fused, exact execution deploys through a standard inference toolchain and transfers to a different VT operator. All TensorRT measurements in this section use an NVIDIA RTX 4070 Ti SUPER (FP32), a different device from the RTX A6000 used above, so TRT numbers should be compared within each table rather than against the PyTorch results.

\paragraph{Deployment in TensorRT.}
TensorRT~\cite{tensorrt} is a widely used inference engine for deploying autonomous-driving perception, and recent BEV methods explicitly target it~\cite{fastbevpp}. We implement FlashBEV as a TensorRT plugin and compare against a Tensorized Sampling-VT plugin under FP32 (Table~\ref{tab:tensorrt}): peak memory stays constant as height bins grow from $Z{=}8$ to $Z{=}32$ while the tensorized baseline grows with $Z$, and FlashBEV remains numerically exact (zero max error).

\begin{table}[t]
\centering
\caption{FlashBEV as a TensorRT plugin vs.\ Tensorized Sampling-VT (FP32), sweeping height bins $Z$. Peak memory (external delta) in MB, latency in ms; $\times$ is Tensorized/FlashBEV. FlashBEV's memory is independent of $Z$ and it remains numerically exact at all $Z$ (zero max error; cf.\ Table~\ref{tab:numerical_diff}).}
\label{tab:tensorrt}
\setlength{\tabcolsep}{4pt}
\begin{tabular}{ccccccc}
\toprule
 & \multicolumn{3}{c}{Peak mem (MB)} & \multicolumn{3}{c}{Latency (ms)} \\
\cmidrule(lr){2-4} \cmidrule(lr){5-7}
$Z$ & Tensorized & FlashBEV & $\times$ & Tensorized & FlashBEV & $\times$ \\
\midrule
8  & 1294 & \textbf{38} & 34  & 5.622  & \textbf{1.128} & 5.0 \\
16 & 2514 & \textbf{38} & 66  & 20.132 & \textbf{2.140} & 9.4 \\
32 & 4994 & \textbf{38} & 131 & 40.193 & \textbf{4.177} & 9.6 \\
\bottomrule
\end{tabular}
\end{table}

\paragraph{Cross-framework applicability: BEVFormer SCA.}
To test whether this execution idea is specific to SimpleBEV-style Sampling-VT, we apply it to the spatial cross-attention (SCA) of BEVFormer~\cite{bevformer}, a deformable-attention view transformation. Our kernel, \textbf{FlashSCA}, performs on-the-fly projection with thread-local recomputation so the per-camera BEV-shaped intermediates are never materialized in HBM; the baseline mirrors the dense per-camera SCA path used in public TensorRT deployments. FlashSCA reproduces BEVFormerSCA's output and gradient up to FP32 round-off ($\max|\Delta|{=}7\times10^{-7}$ forward, $7\times10^{-6}$ backward), so it is an exact, training-compatible replacement requiring no retraining. At BEVFormer-base defaults ($N{=}6$, $C{=}256$, 8 heads, BEV $200{\times}200$, $Z{=}4$, $L{=}4$) we sweep the number of sampling points $P$ in TensorRT (Table~\ref{tab:flashsca}): FlashSCA reduces peak memory by 2.5--3.2$\times$ and latency by 1.7--2.1$\times$, with gains widening in $P$ because BEVFormerSCA materializes a sampling-points tensor of shape $[N{\cdot}X{\cdot}Y, H, L, P, 2]$ that FlashSCA never builds.

\begin{table}[t]
\centering
\caption{FlashSCA vs.\ BEVFormerSCA in TensorRT (FP32) at BEVFormer-base defaults, sweeping sampling points $P$. Peak memory in MB, latency in ms; $\times$ is BEVFormerSCA/FlashSCA.}
\label{tab:flashsca}
\setlength{\tabcolsep}{4pt}
\begin{tabular}{ccccccc}
\toprule
 & \multicolumn{3}{c}{Peak mem (MB)} & \multicolumn{3}{c}{Latency (ms)} \\
\cmidrule(lr){2-4} \cmidrule(lr){5-7}
$P$ & BEVFormerSCA & FlashSCA & $\times$ & BEVFormerSCA & FlashSCA & $\times$ \\
\midrule
8  & 1178 & \textbf{464}  & 2.54 & 8.96  & \textbf{5.27}  & 1.70 \\
16 & 1990 & \textbf{698}  & 2.85 & 15.81 & \textbf{8.16}  & 1.94 \\
32 & 3708 & \textbf{1168} & 3.18 & 29.27 & \textbf{13.90} & 2.11 \\
\bottomrule
\end{tabular}
\end{table}

\section{Conclusion}
\label{sec:conclusion}
Our contribution is an \emph{execution design} for sampling-based view transformation (Sampling-VT): we recast it as a gather--reduction operator and propose \textbf{FlashBEV}, a single fused, IO-aware kernel that preserves exact operator-level equivalence while substantially reducing memory traffic and kernel-launch overhead. FlashBEV reduces VT peak memory by over 37$\times$ and achieves 5.2$\times$ forward speedup (Table~\ref{tab:vt_forward_backward}) with memory independent of height bins; it also deploys in TensorRT and transfers to BEVFormer's SCA (FlashSCA), opening a new design space for scalable BEV perception on commodity and edge devices.

\section*{Acknowledgements}
The authors thank Prof.~Keiji Kimura of Waseda University for insightful discussions.

\bibliographystyle{splncs04}
\bibliography{ref}

\end{document}